\pdfoutput=1

\documentclass[11pt]{article}

\usepackage[]{acl}

\usepackage{times}
\usepackage{latexsym}
\usepackage{listings}
\usepackage{amsmath}

\usepackage[T1]{fontenc}

\usepackage[utf8]{inputenc}

\usepackage{microtype}

\usepackage{inconsolata}

\usepackage{graphicx}

\usepackage{multirow}

\usepackage{array}
\usepackage{longtable}
\usepackage{placeins}
\title{SCORE: \textit{S}ystematic \textit{CO}nsistency and \textit{R}obustness \textit{E}valuation for Large Language Models}

\author{
 \textbf{Grigor Nalbandyan},$\>$
 \textbf{Rima Shahbazyan}, $\>$
 \textbf{Evelina Bakhturina}
\\
NVIDIA\\
\{gnalbandyan, rshahbazyan, ebakhturina\}@nvidia.com
\\
}

\begin{document}
\maketitle
\begin{abstract}
Typical evaluations of Large Language Models (LLMs) report a single metric per dataset, often representing the model's best-case performance under carefully selected settings. Unfortunately, this approach overlooks model robustness and reliability in real-world applications. For instance, simple paraphrasing of prompts on the MMLU-Pro dataset causes accuracy fluctuations of up to 10\%, while reordering answer choices in the AGIEval dataset results in accuracy differences of up to 6.1\%. While some studies discuss issues with LLM robustness, there is no unified or centralized framework for evaluating the robustness of language models. To address this gap and consolidate existing research on model robustness, we present SCORE (\textbf{S}ystematic \textbf{CO}nsistency and \textbf{R}obustness \textbf{E}valuation), a comprehensive framework for non-adversarial evaluation of LLMs. The SCORE framework evaluates models by repeatedly testing them on the same benchmarks in various setups to give a realistic estimate of their accuracy and consistency. 
We release the code\footnote{https://github.com/EleutherAI/lm-evaluation-harness/tree/main/lm\_eval/tasks/score} publicly and start an LLM robustness leaderboard\footnote{https://huggingface.co/spaces/nvidia/llm-robustness-leaderboard} to facilitate further development and research. 

\end{abstract}

\section{Introduction}

The evaluation of Large Language Models (LLMs) typically focuses on a single accuracy metric per dataset, often derived from an optimized setup. This approach provides an incomplete picture of the model capabilities in real-world scenarios.
For an LLM to be trustworthy in practical applications, it must exhibit robustness, i.e., produce consistent responses when the input is rephrased or slightly altered. Consistency is particularly crucial for factual questions in which an objective answer exists. In particular, consistent predictions do not necessarily equate to correct predictions. Given two models with similar accuracy, the one that makes the same incorrect predictions across different setups is arguably preferable.
Recent research has highlighted the limitations of current LLM evaluation practices. \citep{mizrahi2023state, polo2024efficient, alzahrani2024benchmarks} demonstrate the significant impact of simple input perturbations on model performance. \citep{sclar2023quantifying} further underscores the sensitivity of the models to seemingly minor changes in input formatting, such as changing the separator or spacing.
Although robustness analysis is gaining momentum in LLM research, robustness evaluations are often scattered, ad hoc, and difficult to compare between models \cite{dubey2024llama}.

We propose an open evaluation framework SCORE: \textbf{S}ystematic \textbf{CO}nsistency and \textbf{R}obustness \textbf{E}valuation for Large Language Models. SCORE focuses on consistency alongside accuracy to provide a more nuanced understanding of LLM capabilities and facilitate the development of more trustworthy and reliable models. Our contributions are as follows:

\begin{figure}[t]
  \includegraphics[width=220pt, scale=0.4]{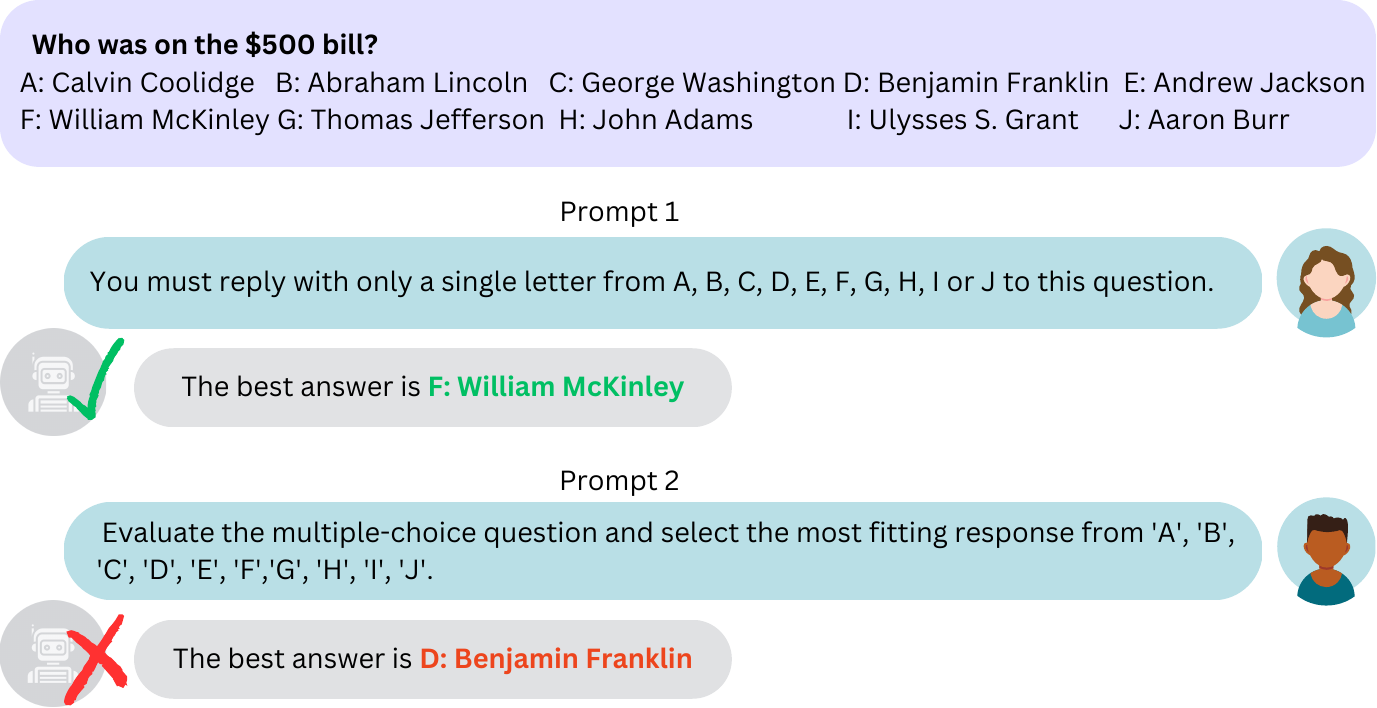}
  \centering
  \caption{Llama-3.1 70B responding inconsistently to an MMLU-Pro question when only prompt is changed.}
  \label{fig:example_1}
\end{figure}

\begin{itemize}
    \item We introduce the SCORE framework, an open and holistic framework that standardizes and unifies the evaluation of the non-adversarial robustness of LLMs.
    \item We investigate the impact of prompt variations, random seed of non-greedy inference, and choice order on model predictions. Our experiments demonstrate that evaluating LLMs across multiple scenarios, considering a range of accuracy values rather than a single metric, and tracking consistency rate provide a more accurate assessment of the model's true capabilities.
    \item We evaluate latest open LLMs to explore the relationship between accuracy and consistency. Our findings reveal that, while these metrics are correlated, higher accuracy or narrow accuracy ranges do not always guarantee better consistency. Furthermore, model size alone is not a reliable indicator of robustness.
\end{itemize}





\section{Related Work}
Open LLM Leaderboard-v2 \cite{open-llm-leaderboard-v2} is a centralized platform for evaluating LLMs in a consistent setup, ensuring fair comparisons. It uses datasets that are both relevant and challenging, but still relies on a single metric evaluation.

PromptBench \cite{zhu2023promptbench, zhu2023promptbench2} focuses on adversarial robustness by providing tools to evaluate models on adversarial prompts — deliberate inputs designed to break their predictions. Although effective, these adversarial attacks could be unrealistic and considerably change the semantics of input samples. PromptBench evaluates models' worst-case performance by estimating how much accuracy degrades under various attacks.

HELM (Holistic Evaluation of Language Models) \cite{liang2023holistic} uses a multi-metric approach to assess the models across various scenarios. However, robustness analysis is limited to character-level perturbations, typos, and a small subset of Contrast Sets \cite{gardner2020evaluating}.

\section{Benchmark}
\begin{figure*}[t]
  \includegraphics[width=\textwidth]{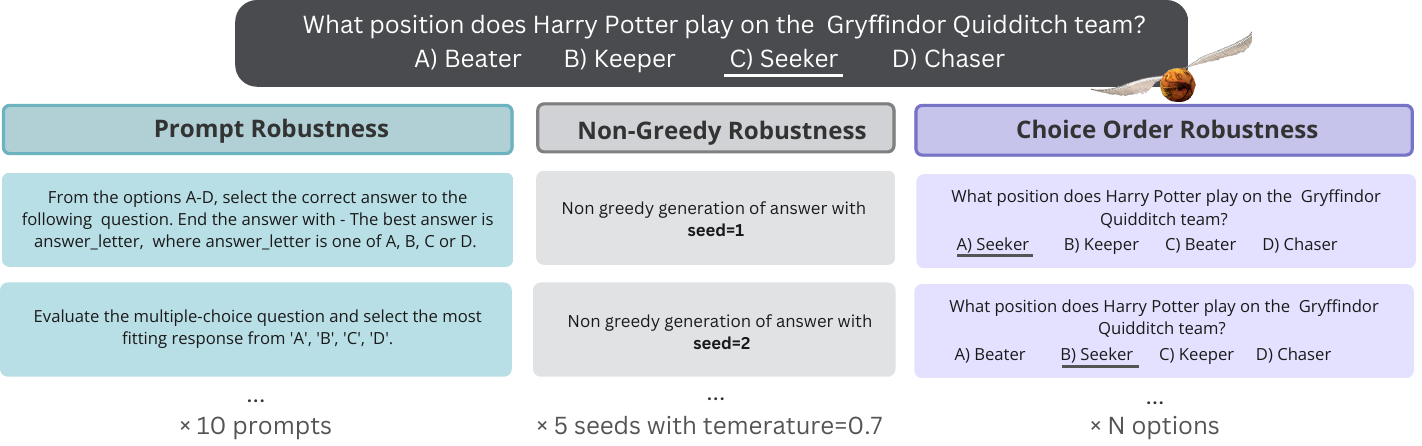}
  \centering
  \caption{Overview of the SCORE robustness tasks. \textit{Prompt Robustness}: This task evaluates multiple-choice question (MCQ) and MATH datasets using ten semantically similar non-adversarial prompts; \textit{Non-Greedy Robustness}: Evaluation is conducted using five random seeds with a fixed prompt, question, and options, with a temperature setting of 0.7; \textit{Choice Order Robustness}: For MCQ datasets, the positions of options are altered while keeping the prompt and question fixed. } 
  \label{fig:example}
\end{figure*}

\subsection{Datasets}
\label{sec:datasets}
To ensure a comprehensive and rigorous evaluation, we employ the following criteria when selecting datasets for our SCORE framework:
\textit{Factuality}: datasets must have objective, verifiable ground truth answers to avoid subjective judgments, such as relying on LLM-as-a-judge evaluation.
\textit{Diversity}: a wide range of topics should be represented to assess model capabilities across various domains.
\textit{Scale}: the datasets should be large enough to ensure the statistical significance of the results.
\textit{Challenging Nature}: the datasets should pose a significant challenge to current open-source LLM models. 
\textit{Minimal Contamination}: as demonstrated by Dubey et al. \cite{dubey2024llama}, widely used benchmarks can be significantly contaminated in the training dataset, which can result in inflated benchmark scores that do not accurately reflect the model's true capabilities. We carefully consider the age and quality of the selected datasets.

Given the substantial computational resources required for multiple evaluations per dataset, we limited our benchmark to the following three open-source datasets that best met our selection criteria - \textbf{MMLU-Pro} \cite{wang2024mmlu}, \textbf{AGIEval} \cite{zhong2023agieval} and \textbf{MATH} \cite{hendrycksmath2021} (see Appendix \ref{sec:datastats} for detailed information on each dataset).

We recognize the limitations of using these datasets, as they do not fully encompass the wide range of use cases that models may encounter in real-world applications. However, they provide a solid foundation for our benchmark. We leave the exploration of additional datasets for future work.

\subsection{Tasks}





\textbf{Prompt Robustness}. The prompt can significantly influence the accuracy and quality of LLM output. Most model evaluation reports contain a single metric corresponding to a tuned and engineered prompt, which maximizes the metric. For a given query, models are expected to get a variety of semantically equivalent prompts. For example, one can think of hundreds of ways to ask a model to solve a mathematical problem. LLMs should to be robust to the changes of prompt formulation and consistent in their answers. A robust LLM will require less prompt engineering as the exact wording of the prompt will not matter for the model. \\
We choose ten prompts and analyze model accuracy and prediction consistency against changing the prompt. The prompts are not adversarial and are not engineered to increase or decrease model accuracy in any way. We include both CoT \cite{wei2022chain} and non-CoT prompts and vary the placement of the question in the prompt to be either in the beginning, in the middle, or at the end of the prompt. For MCQ datasets, prompts ask the model to choose the correct option letter. For MATH, prompts ask the model to solve the problem. 
The full list of prompts can be found in Appendix \ref{sec:prompts}.

\begin{figure*}[t]
  \includegraphics[width=\textwidth]{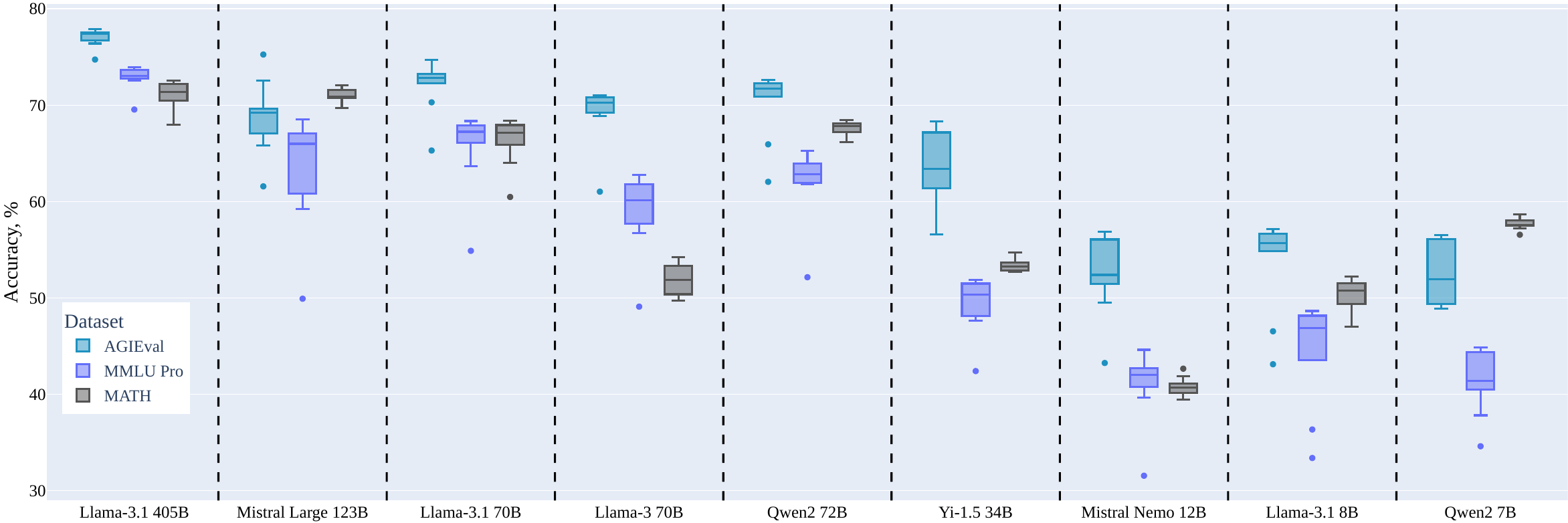}
  \centering
  \caption{Accuracy ranges for Prompt Robustness task on AGIEval, MMLU-Pro and MATH datasets. Evaluation is done using ten distinct prompts (see Appendix \ref{sec:prompts}).}
  \label{fig:prompt}
\end{figure*}

\textbf{Non-Greedy Inference.}
Non-greedy inference is a common technique used to diversify the outputs of LLMs, particularly for queries without objective answers, such as movie recommendations or text paraphrasing. However, for factual questions, the generated answers should remain consistent regardless of the random seed used. The inherent randomness in the answer-generation process can influence the "path" the model takes to arrive at a response. Ideally, the model's underlying distribution should be precise enough that the choice of random seed does not affect the sampling of the next token. \\
We perform non-greedy inference with a temperature of 0.7 and five random seeds across all datasets. Since the datasets are factual, the random seed should have minimal impact on the model's predictions. To reduce computational cost, we use a fixed prompt for the non-greedy task.
\\

\textbf{Choice Order Robustness.}
For multiple-choice question (MCQ) datasets MMLU-Pro and AGIEval, models should choose the correct option letter as an answer, as illustrated in Figure \ref{fig:example}. Both \cite{zheng2023large} and \cite{alzahrani2024benchmarks} demonstrate that even simple changes, such as altering the order of choices, can impact the accuracy of LLMs. These effects may be due to internal model instabilities, biases, or contamination of the test data. Following previous work, we evaluate models against changes in the order of choices for MCQ datasets. We swap the order of options while ensuring the correct answer always corresponds to the same position (all correct answers are A, B, etc.). Changing the order of choices does not alter the input's semantics, so models should ideally remain robust to such minor changes. Although fixing the correct answer to a specific letter could introduce evaluation bias, it also helps identify if the model shows a preference for certain answer options. \\
We expect the model to be resilient to these biases. Unlike prior work, we use generative evaluation instead of log-likelihood, and we analyze prediction consistency along with accuracy.
The same prompt used in the non-greedy evaluation is applied here.

\subsection{Models and Inference Setup}
We include instruct-tuned models from various model families to examine the impact of model size and compare different models of similar scale. All the models included are open-source, and most have been publicly released within the past few months. Specifically, we consider the following models: Llama-3.1 \cite{dubey2024llama} 8B, 70B, 405B, Llama-3 70B\footnote{Llama-3.1-8B-Instruct, Llama-3.1-70B-Instruct, Llama-3.1-405B-Instruct, Meta-Llama-3-70B-Instruct from https://huggingface.co/meta-llama/}, Mistral Nemo 12B, Mistral Large 123B\footnote{Mistral-Nemo-Instruct-2407 and Mistral-Large-Instruct-2407 from https://huggingface.co/mistralai/}, Qwen-2 72B and 7B\footnote{Qwen2-72B-Instruct and Qwen2-7B-Instruct from https://huggingface.co/Qwen/}, and Yi-1.5 34B\footnote{https://huggingface.co/01-ai/Yi-1.5-34B-Chat}. 

We use generative evaluation for all tasks to align with real-world human interactions. This approach, as demonstrated by \citep{wang2024my, lyu2024beyond}, provides a more accurate assessment of LLM performance than log-probability evaluation, particularly for tasks requiring reasoning or computation. The inference setup is explained in more detail in Appendix \ref{inf:app}.

\begin{figure*}[t]
  \includegraphics[width=\textwidth]{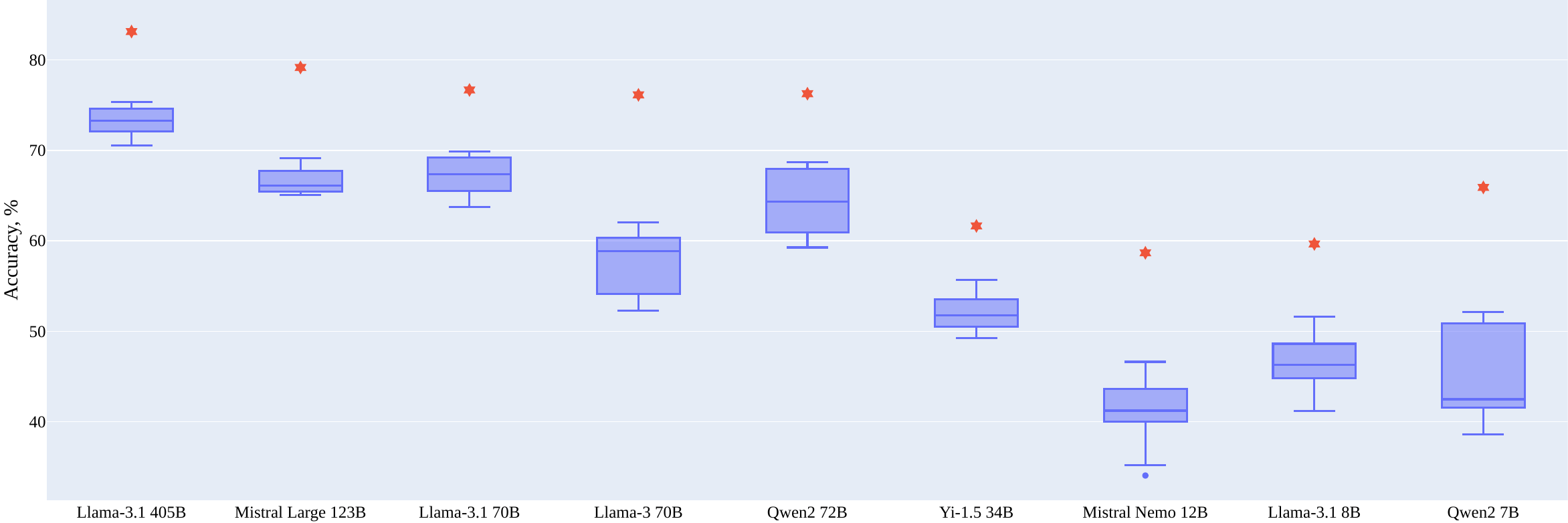}
  \centering
  \caption{Accuracy ranges and Consistency Rate (shown in red) on MMLU-Pro for Choice Order Robustness Task: order of choices is changed while prompt is fixed.}
  \label{fig:order}
\end{figure*}

\subsection{Metrics}

We measure category-level macro accuracy for MMLU-Pro and micro accuracy for AGIEval and MATH, reporting both the mean and the [minimum, maximum] accuracy range. Following \cite{yukun2024improving}, we use \textbf{consistency rate (CR)} to assess the robustness and prediction consistency of LLMs. CR compares all pairs of predictions for a given set of predictions. It is defined as
\begin{equation}
  \label{eq:cr}
  CR = \frac{1}{|Q|} \sum_{Q_k \in Q} \sum_{y_i \in Y_k} \sum_{\substack{y_j \in Y_k \\ j \neq i}} \frac{\text{sim}(y_i, y_j)}{\binom{|Y_k|}{2}}
\end{equation}
where $Q$ is a dataset; $Q_k$ is a single data point; $Y_k$ is the set of predictions for $Q_k$ (e.g. ${|Y_k|=10}$ for prompt robustness); \(y_i\) and \(y_j\) is a pair of predictions for \(Q_k\); $\binom{|Y_k|}{2}$ is the number of all possible prediction pairs and \(sim(y_i, y_j)\) is a similarity function for two predictions.
We extract the final answer from the model's generated text (the choice letter for MCQ and the final answer for MATH) to compute the similarity. For MCQ datasets, we determine the similarity by checking if the two predictions are equal. For MATH, we evaluate the symbolic equivalence between two predictions using the sympy package \cite{meurer2017sympy}. CR does not take the accuracy of individual predictions into account but rather the consistency of the model's responses, e.g., $CR=70\%$ means that 70\% of all prediction pairs for a data point are the same.

\section{Results}
\subsection{Prompt Robustness}

Figure \ref{fig:prompt} illustrates the variation in accuracy across ten prompts for each dataset. There is an outlier prompt, appearing outside the interquartile range of the MMLU-Pro and AGIEval boxplots for all models. This outlier corresponds to the same prompt - \textit{``You must reply with only a single letter from A, B, C, D, E, F, G, H, I or J to this question. Conclude with:\textbackslash n The best answer is answer\_letter where the answer\_letter is a single letter from A to J.\textbackslash n\{QUESTION\}"}. Although the prompt was not deliberately crafted or tuned to reduce accuracy, it causes a significant drop in accuracy and presents a curious phenomenon. We do not include this prompt in the further analysis to avoid making exaggerated claims.
We observe \textbf{no strong correlation between overall accuracy and the spread of accuracy}. Both Mistral models show a variation of 2.3-3.2\% on the MATH dataset, yet their mean accuracy improves significantly from 40.7\% for Mistral 12B to 70.9\% for Mistral Large 123B.
Moreover, models exhibit \textbf{varying accuracy ranges across different datasets}. For example, Yi-1.5 34B accuracy by 2\% on MATH varies, 4.2\% on MMLU-Pro, and 7.6\% on AGIEval. 
It is important to note that \textbf{changes in accuracy do not fully capture prediction stability}, as predictions can vary without affecting the score (e.g., when the model switches from one incorrect prediction to another). 
\textbf{There is a positive correlation between mean accuracy and consistency, but higher accuracy does not always guarantee higher consistency.} For instance, two versions of Llama 70B models - 3 and 3.1 - achieve comparable consistency on the MMLU-Pro dataset (72\% and 70.8\%, respectively). However, Llama-3.1 70B reaches a 6.6\% higher mean accuracy. 
In MCQ datasets, the accuracy varies by 1.5-10.6\% on AGIEval and 1.3-15.2\% on MMLU-Pro, even when excluding the outlier prompt. Across all models, consistency is higher on AGIEval than on MMLU-Pro. This could be attributed to the greater difficulty of MMLU-Pro and the difference in the number of answer choices (up to five for AGIEval versus up to ten for MMLU-Pro). 
\textbf{Accuracy is least sensitive on MATH}, though still varies by 2-7.9\%. \textbf{Prediction consistency on MATH is low for all models} and reaches a maximum of 69.8\%. For Mistral Large 123B, the consistency rate is 69.7\%, and only 60\% of the data points have at least eight equivalent predictions.
Table \ref{tab:prompt} (Appendix \ref{prompt:app}) summarizes the accuracies and consistencies of all models on the prompt robustness task.

\subsection{Choice Order Robustness}
Table \ref{tab:order_choice} (Appendix \ref{ch:app}) summarizes how model predictions and metrics are affected by changes in the order of answer choices. On the MMLU-Pro dataset, accuracy fluctuates between 4\% and 13.5\%, while on AGIEval, the fluctuation is between 2\% and 7.5\% (with up to 29.2\% for Mistral 12B). Figure \ref{fig:order} illustrates the accuracy variance and consistency rate for the choice order robustness task on MMLU-Pro. The wide range of accuracy scores demonstrates why relying on a single number for reporting and model comparison can be misleading. For example, when comparing Llama-3.1 405B and Llama-3.1 70B on MMLU-Pro, accuracy metrics can be very similar (70.5\% vs. 69.5\%) or significantly different (75\% vs 63\%) simply by altering the order of choices. Llama-3.1 405B is more accurate and more consistent on MMLU-Pro dataset.
The Choice Order Robustness experiments align with the findings from the Prompt Robustness tests, demonstrating that \textbf{a higher accuracy does not necessarily imply greater consistency}. For example, while Llama-3.1 70B and Llama-3 70B both achieve a consistency rate of 76\%, the mean accuracy of Llama-3.1 70B is 9.6\% higher. 

\begin{figure*}[h!]
  \includegraphics[width=\textwidth]{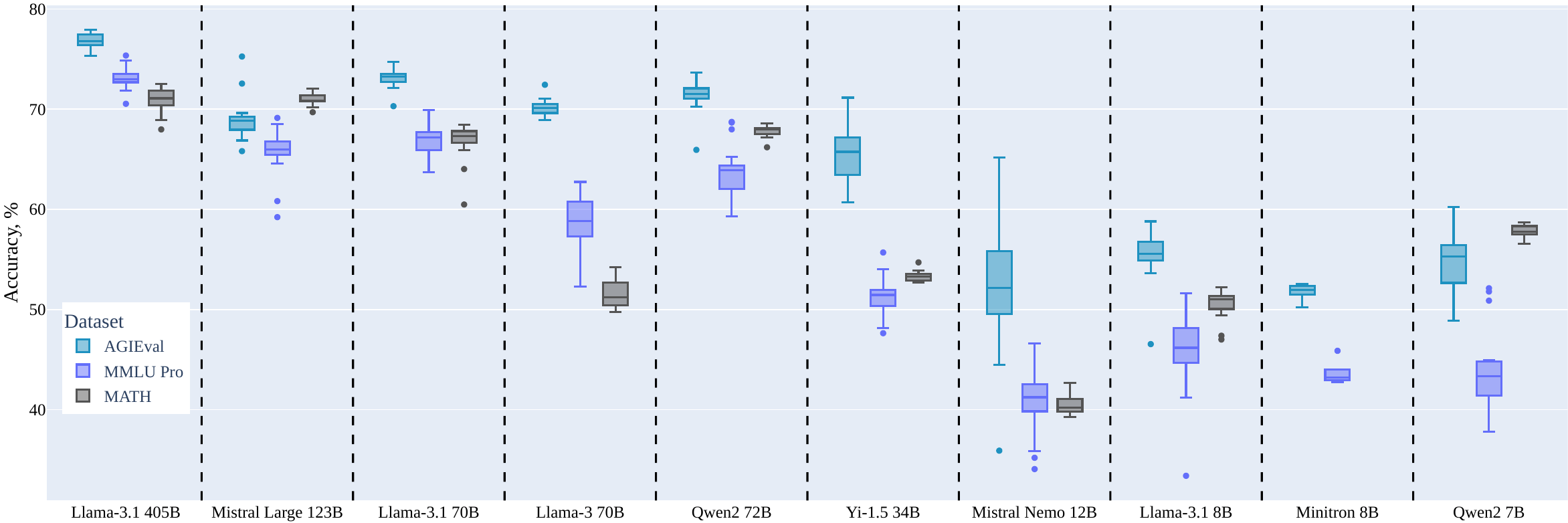}
  \centering
  \caption{Aggregated accuracy ranges for all SCORE robustness tasks and datasets.} 
  \label{fig:combined}
\end{figure*}
\subsection{Non-Greedy Inference}
Table \ref{tab:nongreedy} (Appendix \ref{greedy:app}) aggregates non-greedy inference results across all datasets and models. We observe minimal changes in accuracy, except for Mistral 12B. However, \textbf{despite the stability in accuracy, the consistency rate remains relatively low}, indicating unstable predictions. On the MMLU-Pro, Llama-3.1 405B achieves the highest consistency of 83.3\%, but only 73.4\% of predictions are the same across all seeds. For Llama-3.1 8B, the accuracy varies by 2.32\%, but the consistency rate is only 54.4\%, with 37.9\% of identical predictions across all seeds. Similarly, for MATH, accuracy varies slightly (0.8–3.4\%), but overall consistency is low. The highest consistency rate is 74.6\% for Qwen-2 72B, with 65.7\% of predictions being identical. This \textbf{variability in predictions can be partially attributed to the difficulty of the problems} (see Appendix \ref{sec:topics} for further analysis). For Level 1 problems, 85\% of the predictions are identical between different seeds, while for Level 5 problems, only 29.6\% are consistent. Hence, harder problems mean a more uniform underlying distribution, and changing the seed changes the "path" model takes for a solution. Despite having low accuracy on both datasets, Qwen-2 7B, the smallest model of all, has the highest consistency rate on AGIEval (95.8\%) and MMLU-Pro (92.5\%). 

\subsection{Aggregated Analysis}
Table \ref{tab:final} (Appendix \ref{agg:app}) summarizes the overall consistency rate and accuracy range for each model by averaging the consistency rates across all experiments and aggregating the accuracies (excluding outliers from the MMLU-Pro and AGIEval datasets to avoid exaggerated claims). For instance, aggregated metric for MMLU-Pro includes 24 predictions per data point (nine predictions for Prompt Robustness, ten for Choice Order Robustness, and five for Non-Greedy evaluation). Figure \ref{fig:combined} shows that \textbf{accuracy range varies significantly} depending on the specific model and dataset. For example, Yi-1.5 34B has an accuracy variance of 2\% on MATH but 10.5\% on AGIEval. The variation in metrics can partially be attributed to differences in training data. Llama-3.1 405B is the only model with an accuracy variance below 5\% across all datasets. 
Overall, \textbf{mean accuracy and consistency are correlated}. Across datasets, all models' highest mean accuracy and consistency rate is on AGIEval. 
\textbf{Every model's consistency on MATH is lowest}. This can be partially attributed to the nature of the task, as models must generate the entire answer for the math problem rather than providing a single-letter response, as in standard MCQs. 
\textbf{Model size alone is not a reliable predictor of accuracy and consistency}. For example, Mistral Large 123B is 75\% bigger than Llama-3.1 70B, but CR on MMLU-Pro is 74\% for both, and Llama-3.1's accuracy variance is 6.3\% compared to 9.9\% of Mistral Large. Similarly, Llama-3 70B is almost nine times bigger than Llama-3.1 8B, but the mean accuracy of Llama-3 on MATH is higher by 1.6\%, and consistency is lower by 1.7 points.
The results highlight why \textbf{model comparison based on a single metric could be misleading}. For example, if we focus solely on maximum accuracy—often emphasized in model releases—one might conclude that Yi-1.5 34B performs on par with Llama-3 70B on the AGIEval dataset, despite being half the size. While this is technically true, Yi-1.5 has a wider accuracy range (60.6\% to 71.1\%) compared to Llama-3 70B's range (68.8\% to 72.4\%). Moreover, the consistency rate of Llama-3 70B is 13.2\% higher than Yi-1.5.
Similarly, Mistral Large 123B is 3.2 times smaller than Llama-3.1 405B and its maximum accuracy on AGIEval is only 2.65\% lower than Llama-3.1 405B. However, the accuracy range of Llama-3.1 405B is below 3\% (75. 3\% to 77. 9\%), while the accuracy of Mistral 123B is more sensitive to input changes (65.8\% to 75.2\%). In addition, Llama-3.1 405B has an 11.1\% higher CR.


\section{Conclusion}
Our evaluation demonstrates that relying solely on a single-point evaluation provides an incomplete assessment of the LLM capabilities. We offer a more nuanced and informative understanding of model performance by evaluating models under various conditions and reporting accuracy ranges and consistency rates. Our SCORE framework establishes a foundation for systematic LLM evaluation, facilitating standardized analysis and research of non-adversarial robustness.

\section{Limitations}
The datasets and robustness tests employed in this work may not fully capture the breadth of LLM capabilities. For instance, we rely heavily on MCQ datasets that offer ease of evaluation and factual clarity, and we do not explicitly consider creative tasks such as summarization, where consistency is more subjective. Expanding the scope of evaluation to include additional datasets and robustness tasks would provide an even more complete picture. However, it could also lead to a significant increase in computational costs.
Furthermore, our reliance on publicly available datasets exposes us to the potential risks of data contamination. 

\bibliography{latex/main}

\appendix
\section{Datasets Statistics}
\label{sec:datastats}
\textbf{MATH} \cite{hendrycksmath2021} dataset, which consists of around 5,000 challenging competition-level mathematics problems. Solving these problems requires LLMs to perform multiple reasoning steps to arrive at the correct answer. \\
\begin{table}[!htbp]
\begin{tabular}{cc}
\hline
\textbf{Topic}            & \multicolumn{1}{l}{\textbf{Number of Samples}} \\\hline
Level 1 & 437 \\
Level 2 & 894 \\
Level 3 & 1131 \\
Level 4 & 1214 \\
Level 5 & 1324 \\
\textbf{TOTAL}                              & \textbf{5000}   \\\hline                                
\end{tabular}
\caption{Subset statistics of the MATH dataset (MIT License), categorized by problem difficulty levels.}
\end{table}

\textbf{MMLU-Pro} \cite{wang2024mmlu} is an enhanced version of MMLU \cite{hendrycks2020measuring}, a widely used multiple-choice benchmark for evaluating the core knowledge and reasoning abilities of LLMs. MMLU-Pro increases the number of answer choices from 4 to 10, incorporates more reasoning-based questions, and removes incorrect or outdated content. It includes 12,032 questions across 14 subjects, covering a broad range of topics such as natural sciences, business, engineering, and law. Overall, MMLU-Pro provides a higher quality and more challenging assessment than its predecessor.
\begin{table}[!htbp]
\begin{tabular}{cc}
\hline
\textbf{Topic}            & \multicolumn{1}{l}{\textbf{Number of Samples}} \\\hline
biology          & 717                                   \\
business         & 789                                   \\
chemistry        & 1132                                  \\
computer science & 410                                   \\
economics        & 844                                   \\
engineering      & 969                                   \\
health           & 818                                   \\
history          & 381                                   \\
law              & 1101                                  \\
math             & 1351                                  \\
other            & 924                                   \\
philosophy       & 499                                   \\
physics          & 1299                                  \\
psychology       & 798                                   \\
\textbf{TOTAL}            & \textbf{12032}   \\\hline 
\end{tabular}
\caption{Subset statistics of the MMLU-Pro dataset (Apache License Version 2.0), categorized by subject.}
\end{table}
\newpage
\textbf{AGIEval} \cite{zhong2023agieval} is a multiple-choice dataset derived from standardized exams such as SAT and LSAT. It tests models' abilities in reading comprehension, reasoning, and mathematics. For our analysis, we selected SAT-English, SAT-Math, LSAT-Analytics, LSAT-Logic, LSAT-Reading, LogiQA-En, and AQuA-RAT (GRE, GMAT) subsets comprising 2340 datapoints.\\

\begin{table}[!htbp]
\begin{tabular}{cc}
\hline
\textbf{Topic}            & \multicolumn{1}{l}{\textbf{Number of Samples}} \\\hline
aqua\_rat                          & 254                                            \\
logiqa\_en                         & 651                                            \\
lsat\_ar                           & 230                                            \\
lsat\_lr                           & 510                                            \\
lsat\_rc                           & 269                                            \\
sat\_en                            & 206                                            \\
sat\_math                          & 220                                            \\
\textbf{TOTAL}                              & \textbf{2340} \\\hline                                            \\
\end{tabular}
\caption{Subset statistics of the AGIEval dataset (MIT License).}
\end{table}

\newpage
\section{Inference Setup}
\label{inf:app}
While generative evaluation incurs higher computational costs than other methods, the additional expense is negligible compared to the overall training costs.
For each model and dataset, we generate 1024 tokens. We found that over 95\% of the datasets can be answered within this token limit, with models occasionally getting stuck in repetitive loops that require generating more tokens. Generating additional tokens beyond this limit yields diminishing returns in metrics while significantly increasing computational costs. To simulate average user behaviour, we conduct all evaluations in a 0-shot setting, without providing any few-shot examples. 
Model predictions are extracted by parsing the generated text. For MATH problems, we instruct the model to format its answer within \$\symbol{92}\symbol{92}boxed\{\{answer\}\}\$ to extract prediction easily and verify its symbolic equivalence with the ground truth using the sympy package \cite{meurer2017sympy}. In the case of MCQ, the model is prompted to conclude with \textit{The best answer is answer\_letter}, and the corresponding letter is extracted from the output. While more complex post-processing might improve metrics by addressing cases where models deviate from instructions, we avoid such techniques to maintain a model-independent parsing logic and ensure that models follow the given prompts. \\
We convert models to TRT-LLM\footnote{https://github.com/NVIDIA/TensorRT-LLM} for evaluation. We have used two NVIDIA A100 80GB nodes for Llama-3.1 405B evaluation and a single node for the rest of the models. For the SCORE evaluation, we conducted a series of robustness evaluations for each model: 25 evaluations on the MMLU-Pro dataset (ten predictions for Prompt Robustness, ten for Choice Order Robustness, and five for Non-greedy evaluation), 19 on the AGIEval dataset (ten predictions for Prompt Robustness, four for Choice Order Robustness, and five for Non-greedy evaluation), and 15 on the MATH dataset (ten predictions for Prompt Robustness, and five for Non-greedy evaluation). The specific computational requirements for each evaluation varied depending on the model size, dataset size, and the model's verbosity in generating answers.
\newpage
\section{Prompt Robustness Results}
\label{prompt:app}
\begin{table}[h]
\resizebox{\columnwidth}{!}{
\begin{tabular}{ccc}
\hline
\multirow{2}{*}{\textbf{Model}} & \textbf{Accuracy, \%}   & \multirow{2}{*}{\textbf{CR}} \\
                       & \textbf{Mean {[}Min, Max{]}}   &                          \\\hline
\multicolumn{3}{c}{\textbf{AGIEval}} \\
Llama-3.1 405B         & 77.0 {[}74.7, 77.9{]}       & 86.1 \\
Mistral Large 123B     & 68.8 {[}61.5, 75.2{]}       & 74.3 \\
Qwen-2 72B             & 70.2 {[}62.0, 72.5{]}       & 80.5 \\
Llama-3.1 70B          & 72.0 {[}65.3, 74.7{]}       & 80.5 \\
Llama-3 70B            & 69.2 {[}61.0, 71.0{]}       & 80.5 \\
Yi-1.5 34B             & 63.6 {[}56.5, 68.3{]}       & 66.4 \\
Mistral Nemo 12B       & 52.4 {[}43.2, 56.8{]}       & 58.9 \\
Llama-3.1 8B           & 53.9 {[}43.1, 57.1{]}       & 59.7 \\
Qwen-2 7B              & 52.4 {[}48.8, 56.4{]}       & 61.5
\\\hline
\multicolumn{3}{c}{\textbf{MMLU-Pro}} \\
Llama-3.1 405B         & 72.8 {[}69.5, 73.9{]}       & 79.8 \\
Mistral Large 123B     & 63.6 {[}49.9, 68.5{]}       & 70.2 \\
Qwen-2 72B             & 62.1 {[}52.5, 65.2{]}       & 72.2 \\
Llama-3.1 70B          & 65.7 {[}54.8, 68.3{]}       & 72.0 \\
Llama-3 70B            & 59.1 {[}49.1, 62.7{]}       & 70.8 \\
Yi-1.5 34B             & 49.4 {[}42.4, 51.9{]}       & 53.3 \\
Mistral Nemo 12B       & 41.0 {[}31.5, 44.6{]}       & 46.7 \\
Llama-3.1 8B           & 44.4 {[}33.3, 48.6{]}       & 47.9 \\
Qwen-2 7B              & 41.3 {[}34.6, 44.8{]}       & 49.1 \\
\hline
\multicolumn{3}{c}{\textbf{MATH}} \\
Llama-3.1 405B         & 71.0 {[}67.9, 72.5{]}       & 69.8 \\
Mistral Large 123B     & 70.9 {[}69.7, 72.0{]}       & 69.7 \\
Qwen-2 72B             & 67.6 {[}52.1, 65.2{]}       & 72.2 \\
Llama-3.1 70B          & 66.3 {[}60.4, 68.4{]}       & 64.6 \\
Llama-3 70B            & 51.8 {[}49.7, 54.2{]}       & 50.1 \\
Yi-1.5 34B             & 53.3 {[}52.7, 54.7{]}       & 48.0 \\
Mistral Nemo 12B       & 40.7 {[}39.4, 42.6{]}       & 36.9 \\
Llama-3.1 8B           & 50.2 {[}47.0, 52.2{]}       & 46.0 \\
Qwen-2 7B              & 57.6 {[}56.5, 58.6{]}       & 58.3 \\
\hline
\end{tabular}}
\caption{Accuracy ranges and consistency rates (CR) on Prompt Robustness task: the evaluation is conducted using 10 prompts, while keeping the context fixed.}
\label{tab:prompt}
\end{table}
\clearpage

\section{Choice Order Robustness Results}
\label{ch:app}

\begin{table}[h]
\resizebox{\columnwidth}{!}{
\begin{tabular}{ccc}
\hline
\multirow{2}{*}{\textbf{Model}} & \textbf{Accuracy, \%}   & \multirow{2}{*}{\textbf{CR}} \\
                       & \textbf{Mean {[}Min, Max{]}}   &                          \\\hline
\multicolumn{3}{c}{\textbf{AGIEval}}                                       \\
Llama-3.1 405B         & 76.5 {[}75.3, 77.3{]}       & 88.5                \\
Mistral Large 123B     & 68.2 {[}66.8, 68.8{]}       & 78                  \\
Qwen-2 72B             & 71.7 {[}70.2, 73.6{]}       & 80.8                \\
Llama-3.1 70B          & 73.6 {[}72.1, 74.7{]}       & 82.5                \\
Llama-3 70B            & 70.3 {[}69.1, 72.4{]}       & 84.1                \\
Yi-1.5 34B             & 68.1 {[}65.0, 71.1{]}       & 71.8                \\
Mistral Nemo 12B       & 51.6 {[}35.9, 65.1{]}       & 61.2                \\
Llama-3.1 8B           & 56.2 {[}53.8, 58.8{]}       & 67.2                \\
Qwen-2 7B              & 55.6 {[}52.6, 60.2{]}       & 72.3                \\
\hline
\multicolumn{3}{c}{\textbf{MMLU-Pro}}                                      \\
Llama-3.1 405B         & 73.1 {[}70.5, 75.3{]}       & 83.1                \\
Mistral Large 123B     & 66.4 {[}65.0, 69.1{]}       & 79.1                \\
Qwen-2 72B             & 64.0 {[}59.2, 68.7{]}       & 76.2                \\
Llama-3.1 70B          & 67.0 {[}63.7, 69.9{]}       & 76.6                \\
Llama-3 70B            & 57.5 {[}52.2, 62.0{]}       & 76.1                \\
Yi-1.5 34B             & 52.0 {[}49.2, 55.6{]}       & 61.6                \\
Mistral Nemo 12B       & 40.8 {[}34.0, 46.6{]}       & 58.6                \\
Llama-3.1 8B           & 46.2 {[}41.1, 51.6{]}       & 59.6                \\
Qwen-2 7B              & 44.4 {[}38.5, 52.2{]}       & 65.9                \\
\hline                    
\end{tabular}}
\caption{Accuracy and consistency rates (CR) for Choice Order Robustness task: order of choices is changed while prompt is fixed.}
\label{tab:order_choice}
\end{table}

\newpage
\section{Non Greedy Results}
\label{greedy:app}
\begin{table}[h]
\resizebox{\columnwidth}{!}{
\begin{tabular}{ccc}
\hline
\multirow{2}{*}{\textbf{Model}} & \textbf{Accuracy, \%}   & \multirow{2}{*}{\textbf{CR}} \\
                       & \textbf{Mean {[}Min, Max{]}}   &                          \\\hline
\multicolumn{3}{c}{\textbf{AGIEval}}                                       \\
Llama-3.1 405B         & 76.4 {[}75.9, 76.7{]}       & 91.1                \\
Mistral Large 123B     & 68.3 {[}67.4, 69.1{]}       & 78.1                \\
Qwen-2 72B             & 71.2 {[}70.8, 71.7{]}       & 91.5                \\
Llama-3.1 70B          & 73.2 {[}72.8, 73.5{]}       & 85.2                \\
Llama-3 70B            & 70.0 {[}69.7, 70.2{]}       & 89.4                \\
Yi-1.5 34B             & 66.1 {[}65.4, 67.0{]}       & 74.9                \\
Mistral Nemo 12B       & 49.4 {[}44.4, 52.8{]}       & 53.9                \\
Llama-3.1 8B           & 54.9 {[}53.6, 56.6{]}       & 66.6                \\
Qwen-2 7B              & 56.2 {[}55.3, 56.5{]}       & 95.8                \\
\hline
\multicolumn{3}{c}{\textbf{MMLU-Pro}}                                      \\
Llama-3.1 405B         & 72.7 {[}72.6, 72.9{]}       & 83.3                \\
Mistral Large 123B     & 65.8 {[}65.3, 66.0{]}       & 76.0                \\
Qwen-2 72B             & 63.7 {[}63.7, 64.0{]}       & 86.9                \\
Llama-3.1 70B          & 66.5 {[}65.6, 67.1{]}       & 74.8                \\
Llama-3 70B            & 57.4 {[}57.2, 57.6{]}       & 79.8                \\
Yi-1.5 34B             & 51.5 {[}51.4, 51.9{]}       & 63.0                \\
Mistral Nemo 12B       & 39.0 {[}35.8, 40.9{]}       & 47.2                \\
Llama-3.1 8B           & 45.1 {[}44.1, 46.2{]}       & 54.4                \\
Qwen-2 7B              & 44.6 {[}44.4, 44.9{]}       & 92.5                \\
\hline
\multicolumn{3}{c}{\textbf{MATH}}                                          \\
Llama-3.1 405B         & 70.6 {[}70.2, 71.1{]}       & 68.0                \\
Mistral Large 123B     & 70.9 {[}70.5, 71.4{]}       & 68.6                \\
Qwen-2 72B             & 68.0 {[}67.4, 68.5{]}       & 74.6                \\
Llama-3.1 70B          & 67.3 {[}66.7, 68.1{]}       & 65.0                \\
Llama-3 70B            & 50.8 {[}50.0, 51.6{]}       & 48.6                \\
Yi-1.5 34B             & 53.2 {[}52.8, 53.6{]}       & 48.0                \\
Mistral Nemo 12B       & 40.1 {[}39.2, 42.6{]}       & 33.7                \\
Llama-3.1 8B           & 50.8 {[}49.6, 51.9{]}       & 45.5                \\    
Qwen-2 7B              & 58.1 {[}57.2, 58.6{]}       & 68.3                \\ 
\hline                    
\end{tabular}}
\caption{Accuracy ranges and consistency rates (CR) for Non-Greedy Robustness tasks: models evaluated on five random seeds with temperature set to 0.7.}
\label{tab:nongreedy}
\end{table}

\newpage
\section{Aggregated  Results}
\label{agg:app}
\begin{table}[h]
\resizebox{\columnwidth}{!}{
\begin{tabular}{ccc}
\hline
\multirow{2}{*}{\textbf{Model}} & \textbf{Accuracy, \%}   & \multirow{2}{*}{\textbf{CR}} \\
                       & \textbf{Mean {[}Min, Max{]}}   &                          \\\hline
\multicolumn{3}{c}{\textbf{AGIEval}}                                       \\
Llama-3.1 405B         & 77.0 {[}75.3, 77.9{]}       & 87.3                \\
Mistral Large 123B     & 69.2 {[}65.8, 75.2{]}       & 76.2                \\
Qwen-2 72B             & 71.3 {[}65.9, 73.6{]}       & 80.7                \\
Llama-3.1 70B          & 73.0 {[}70.3, 74.7{]}       & 81.7                \\
Llama-3 70B            & 70.2 {[}68.8, 72.4{]}       & 82.3                \\
Yi-1.5 34B             & 65.6 {[}60.6, 71.1{]}       & 69.1                \\
Mistral Nemo 12B       & 52.9 {[}35.9, 65.1{]}       & 60.0                \\
Llama-3.1 8B           & 55.3 {[}46.5, 58.8{]}       & 63.3                \\ 
Qwen-2 7B              & 53.6 {[}48.8, 60.2{]}       & 66.9                \\ 
\hline
\multicolumn{3}{c}{\textbf{MMLU-Pro}}                                      \\
Llama-3.1 405B         & 73.1 {[}70.5, 75.3{]}       & 81.5                \\
Mistral Large 123B     & 65.8 {[}59.2, 69.1{]}       & 74.7                \\
Qwen-2 72B             & 63.6 {[}59.2, 68.7{]}       & 74.2                \\
Llama-3.1 70B          & 67.0 {[}63.6, 69.9{]}       & 74.3                \\
Llama-3 70B            & 58.8 {[}52.2, 62.7{]}       & 73.5                \\
Yi-1.5 34B             & 51.2 {[}47.6, 55.6{]}       & 57.4                \\
Mistral Nemo 12B       & 41.4 {[}34.0, 46.6{]}       & 52.6                \\
Llama-3.1 8B           & 45.8 {[}33.3, 51.6{]}       & 53.8                \\ 
Qwen-2 7B              & 43.3 {[}37.8, 52.1{]}       & 57.5                \\ 
\hline
\multicolumn{3}{c}{\textbf{MATH}}                                          \\
Llama-3.1 405B         & 71.0 {[}67.9, 72.5{]}       & 71.1                \\
Mistral Large 123B     & 70.9 {[}69.7, 72.0{]}       & 70.6                \\
Qwen-2 72B             & 67.6 {[}66.2, 68.4{]}       & 68.6                \\
Llama-3.1 70B          & 66.3 {[}60.4, 68.4{]}       & 67.0                \\
Llama-3 70B            & 51.8 {[}49.7, 54.2{]}       & 50.4                \\
Yi-1.5 34B             & 53.3 {[}52.7, 54.7{]}       & 49.4                \\
Mistral Nemo 12B       & 40.7 {[}39.2, 42.6{]}       & 38.2                \\
Llama-3.1 8B           & 50.2 {[}47.0, 52.2{]}       & 52.1                \\ 
Qwen-2 7B              & 57.6 {[}56.5, 58.6{]}       & 61.0                \\ 
\hline
\end{tabular}}
\caption{Accuracy ranges and consistency rates (CR) aggregated across Prompt Robustness, Choice Order Robustness and random seed variation for Non-greedy inference.}
\label{tab:final}
\end{table}

\clearpage
\onecolumn 

\section{Prompts}
\label{sec:prompts}

\subsection{MMLU-Pro Prompts}

\begin{table}[h!]
\begin{tabular}{l}
\begin{minipage}[H]{0.99\textwidth}
----------------------------------------------------------------------------------------------------------------------------\\
    \{\}
    Examine the question and choose the correct answer from the options 'A', 'B', 'C', 'D', 'E', 'F', 'G', 'H', 'I' or 'J'. End your answer with:
    The best answer is [the\_answer\_letter].
    where the [the\_answer\_letter] is a letter from A to J.\\
----------------------------------------------------------------------------------------------------------------------------\\
    \{\}
    Answer the multiple-choice question about {task} by selecting the correct option from A to J. Always conclude with 'The best answer is (answer\_letter)' where the (answer\_letter) is one of A, B, C, D, E, F, G, H, I, J.\\
    ----------------------------------------------------------------------------------------------------------------------------\\
    You must reply with only a single letter from A, B, C, D, E, F, G, H, I or J to this question. Conclude with:
    The best answer is answer\_letter where the answer\_letter is a single letter from A to J.
    \{\}\\
    ----------------------------------------------------------------------------------------------------------------------------\\
    From the options A-J, select the correct answer to the following question. End the answer with - The best answer is answer\_letter, where answer\_letter is one of A, B, C, D, E, F, G, H, I, or J.
    Question: \{\}\\
    ----------------------------------------------------------------------------------------------------------------------------\\
    For the multiple-choice question related to {task}, which option (A-J) is correct?.
    
    Question:\{\}
    End the answer with the following:
    The best answer is (the\_answer\_letter) where the (the\_answer\_letter) is one of 'A', 'B', 'C', 'D', 'E', 'F', 'G', 'H', 'I' or 'J'. \\ \textbf{*Used as the fixed prompt for Choice Order and Non-greedy Robustness tasks} \\
    ----------------------------------------------------------------------------------------------------------------------------\\
    Evaluate the multiple-choice question and select the most fitting response from 'A', 'B', 'C', 'D', 'E', 'F', 'G', 'H', 'I', 'J'. 
    Question:\{\}
    Always conclude with:
    The best answer is [the\_answer\_letter].
    where the [the\_answer\_letter] is one of A, B, C, D, E, F, G, H, I or J.\\
    ----------------------------------------------------------------------------------------------------------------------------\\
    Answer to the following question about {task} by selecting the correct option A, B, C, D, E, F, G, H, I or J. \{\}
    The answer should end with:
    The best answer is [the\_answer\_letter] where [the\_answer\_letter] is one of letters A to J. Let's think step by step.\\
    ----------------------------------------------------------------------------------------------------------------------------\\
    Select the correct answer from the options 'A', 'B', 'C', 'D', 'E', 'F', 'G', 'H', 'I','J' for the question provided below. Conclude by stating: The best answer is answer\_letter where answer\_letter is one of 'A', 'B', 'C', 'D', 'E', 'F', 'G', 'H', 'I' or 'J'.
    Question: \{\}
    Let's think step by step.\\
    ----------------------------------------------------------------------------------------------------------------------------\\
   \{\}
   For this question about {task} with 10 possible answers A, B, C, D, E, F, G, H, I, J, choose the one that answers the question. If the problem is simple or straightforward, just provide the answer. If the answer is more complex, use a step-by-step approach and for each step briefly explain your reasoning. Always conclude with 'The best answer is (answer\_letter)' where the (answer\_letter) is one of 'A', 'B', 'C', 'D', 'E', 'F', 'G', 'H', 'I','J'. Let's think step by step.\\
    ----------------------------------------------------------------------------------------------------------------------------\\
    Read the question and options below, then determine the correct answer choice (A-J)
    Question: \{\}
    
    For simple questions, provide a quick answer. For complicated ones, think step by step, break down the question into smaller problems and reach to a conclusion
    End your answer by stating:
    The best answer is [the\_answer\_letter].
    where [the\_answer\_letter] is one of A, B, C, D, E, F, G, H, I, or J. \\
    ----------------------------------------------------------------------------------------------------------------------------\\

\end{minipage}\tabularnewline
\end{tabular}
\caption{Prompts used to evaluate LLMs on the MMLU-Pro dataset for SCORE tasks. \textit{\{\}} denotes a placeholder for an input query.}
\end{table}

\clearpage

\subsection{AGIEval Prompts}

\begin{table}[h!]
\begin{tabular}{l}
\begin{minipage}[H]{0.99\textwidth}
----------------------------------------------------------------------------------------------------------------------------\\
    \{\}
    Examine the question and choose the correct answer from the options 'A', 'B', 'C', 'D' or 'E'. End your answer with:
    The best answer is [the\_answer\_letter].
    where the [the\_answer\_letter] is a letter from A to E.\\
    ----------------------------------------------------------------------------------------------------------------------------\\
    \{\}
    Answer the multiple-choice question by selecting the correct option from A to E. Always conclude with 'The best answer is (answer\_letter)' where the (answer\_letter) is one of A, B, C, D, E.\\
    ----------------------------------------------------------------------------------------------------------------------------\\
    You must reply with only a single letter from A, B, C, D or E to this question. Conclude with:
    The best answer is answer\_letter where the answer\_letter is a single letter from A to E.
    \{\}\\
    ----------------------------------------------------------------------------------------------------------------------------\\
    From the options A-E, select the correct answer to the following question. End the answer with - The best answer is answer\_letter, where answer\_letter is one of A, B, C, D or E.
    Question: \{\}\\
    ----------------------------------------------------------------------------------------------------------------------------\\
    For the multiple-choice question, which option (A-E) is correct?.
    
    Question:\{\}
    End the answer with the following:
    The best answer is (the\_answer\_letter) where the (the\_answer\_letter) is one of 'A', 'B', 'C', 'D' or 'E'.\\
    ----------------------------------------------------------------------------------------------------------------------------\\
    Evaluate the multiple-choice question and select the most fitting response from 'A', 'B', 'C', 'D', 'E'. 
    Question:\{\}
    Always conclude with:
    The best answer is [the\_answer\_letter].
    where the [the\_answer\_letter] is one of A, B, C, D or E. \\ \textbf{*Used as a fixed prompt for Choice Order and Non-greedy Robustness tasks} \\
    ----------------------------------------------------------------------------------------------------------------------------\\
    Answer to the following question by selecting the correct option A, B, C, D or E. \{\}
    The answer should end with:
    The best answer is [the\_answer\_letter] where [the\_answer\_letter] is one of letters A to E. Let's think step by step.\\
    ----------------------------------------------------------------------------------------------------------------------------\\
    Select the correct answer from the options 'A', 'B', 'C', 'D', 'E' for the question provided below. Conclude by stating: The best answer is answer\_letter where answer\_letter is one of 'A', 'B', 'C', 'D' or 'E'.
    Question: \{\}
    Let's think step by step.\\
    ----------------------------------------------------------------------------------------------------------------------------\\
    \{\}
    For this question with 10 possible answers A, B, C, D, E, choose the one that answers the question. If the problem is simple or straightforward, just provide the answer. If the answer is more complex, use a step-by-step approach and for each step briefly explain your reasoning. Always conclude with 'The best answer is (answer\_letter)' where the (answer\_letter) is one of 'A', 'B', 'C', 'D', 'E'. Let's think step by step.\\
    ----------------------------------------------------------------------------------------------------------------------------\\
    Read the question and options below, then determine the correct answer choice (A-E)
    Question: \{\}
    
    For simple questions, provide a quick answer. For complicated ones, think step by step, break down the question into smaller problems and reach to a conclusion
    End your answer by stating:
    The best answer is [the\_answer\_letter].
    where [the\_answer\_letter] is one of A, B, C, D or E.\\
    ----------------------------------------------------------------------------------------------------------------------------\\

\end{minipage}\tabularnewline
\end{tabular}
\caption{Prompts used to evaluate LLMs on the AGIEval dataset for SCORE tasks. \textit{\{\}} denotes a placeholder for an input query.}
\end{table}

\clearpage

\subsection{MATH Prompts}

\begin{table}[h!]
\begin{tabular}{l}
\begin{minipage}[H]{0.99\textwidth}
----------------------------------------------------------------------------------------------------------------------------\\
    Solve this math problem. Your answer should end with 'The final answer is: \$\symbol{92}\symbol{92}boxed\{\{answer\}\}\$ where [answer] is just the final number or expression that solves the problem
    Problem: \{question\}\\
    ----------------------------------------------------------------------------------------------------------------------------\\
    \{question\}
    Please solve this math problem efficiently. Finish with: The final answer is: \$\symbol{92}\symbol{92}boxed\{\{answer\}\}\$  where [answer] is just the final number or expression that solves the problem.\\
    ----------------------------------------------------------------------------------------------------------------------------\\
    Find the answer to the following math question. Conclude with: 'The final answer is: \$\symbol{92}\symbol{92}boxed\{\{answer\}\}\$ '
    where [answer] is just the final number or expression that solves the problem
    Problem: \{question\}\\
    ----------------------------------------------------------------------------------------------------------------------------\\
    \{question\}
    Find the solution to this math problem. Your answer should end with - The final answer is: \$\symbol{92}\symbol{92}boxed\{\{answer\}\}\$ 
    where [answer] is just the final number or expression that solves the problem.\\
    ----------------------------------------------------------------------------------------------------------------------------\\
    Analyze and solve the math task.
    Problem: \{question\}
    End the answer with:
    The final answer is: \$\symbol{92}\symbol{92}boxed\{\{answer\}\}\$  where [answer] is just the final number or expression that solves the problem.\\
    ----------------------------------------------------------------------------------------------------------------------------\\
    Calculate the answer to this math problem
    Problem: \{question\}
    Conclude your answer with:
    The final answer is: \$\symbol{92}\symbol{92}boxed\{\{answer\}\}\$ 
    where [answer] is just the final number or expression that solves the problem.\\
    \textbf{*Used as a fixed prompt for Choice Order and Non-greedy Robustness tasks}\\
    ----------------------------------------------------------------------------------------------------------------------------\\
    \{question\}
    Solve the following math problem
    Show each step of your solution
    Conclude with:
    The final answer is: \$\symbol{92}\symbol{92}boxed\{\{answer\}\}\$\textbackslash nwhere [answer] is just the final number or expression that solves the problem
    Lets think step by step\\
    ----------------------------------------------------------------------------------------------------------------------------\\
    Efficiently solve the following math challenge. Explain your approach step-by-step
    The answer should end with: The final answer is: \$\symbol{92}\symbol{92}boxed\{\{answer\}\}\$ 
    where [answer] is just the final number or expression that solves the problem
    Problem: \{question\}
    Lets think step by step\\
    ----------------------------------------------------------------------------------------------------------------------------\\
    Please solve the math problem. For simple problems offer a quick solution with minimal details. For more challenging problems, explain your approach step-by-step. Finish with
    The final answer is: \$\symbol{92}\symbol{92}boxed\{\{answer\}\}\$ .
    where [answer] is just the final number or expression that solves the problem.
    Problem: \{question\}
    Lets think step by step.\\
    ----------------------------------------------------------------------------------------------------------------------------\\
    You should solve this math problem.
    If the problem is easy, provide a brief solution with little explanation.
    For more difficult problems, follow this structured format\\
    \#\# Step 1: [Brief description]
    [Simple explanation and calculations]
    
    \#\# Step 2: [Brief description]
    [Simple explanation and calculations]
    
    Repeat steps until your reach a solution
    
    Problem: \{question\}
    End with:
    The final answer is: \$\symbol{92}\symbol{92}boxed\{\{answer\}\}\$ 
    where [answer] is just the final number or expression that solves the problem.\\
    ----------------------------------------------------------------------------------------------------------------------------\\

\end{minipage}\tabularnewline
\end{tabular}
\caption{Prompts used to evaluate LLMs on the MATH dataset for SCORE tasks. \textit{\{question\}} denotes a placeholder for an input query.}
\end{table}

\clearpage

\FloatBarrier
\section{Per Topic Analysis}
\label{sec:topics}
\begin{figure}[!htbp]
  \includegraphics[width=0.92\textwidth]{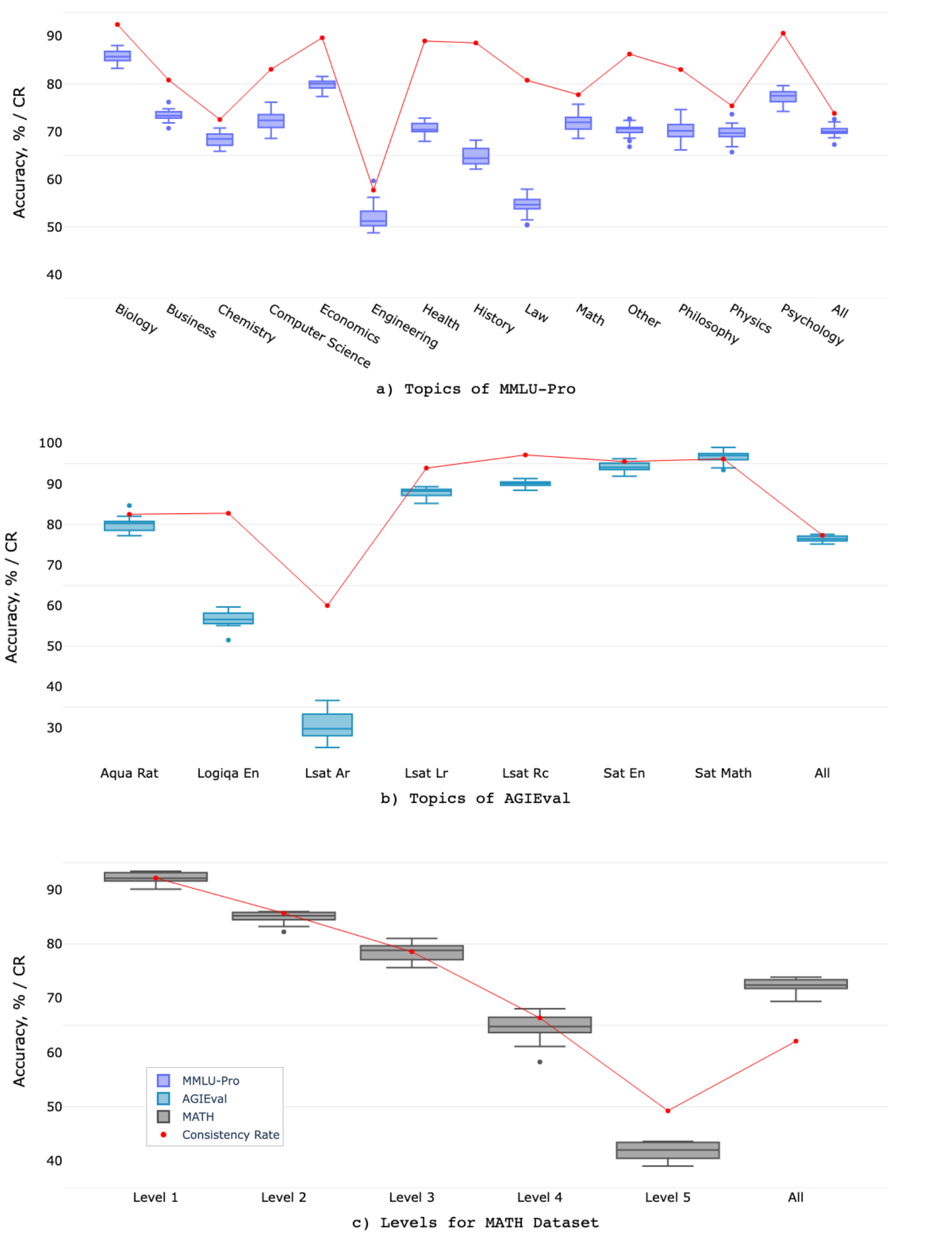}
  \centering
  \caption{Accuracy ranges and consistency rates (CR) for Llama-3.1 405B model across three datasets \textit{a)} MMLU-Pro \textbf{b)} AGIEval \textit{c)} MATH. Each plot represents values across corresponding to specific topics or areas of the dataset (see Appendix \ref{sec:datastats} for details). \textit{"All"} - indicates the accuracy and consistency values for the entire dataset. \\
  For MMLU-Pro, consistency is not uniformly distributed, and accuracy varies between 3.8\% and 10.9\%. There are tasks with \textbf{same consistency but varying accuracy} (e.g., health vs. history) and \textbf{same accuracy but varying consistency} (e.g., physics vs. other). For AGIEval, the accuracy variance across subsets ranges from a maximum of 1\% on LSAT-AR to a minimum of 2.3\% on SAT-EN. In the case of MATH, the trend is clear: as question complexity increases, accuracy decreases, consistency declines, and variance grows.}
  \label{fig:per_task}
\end{figure}

\end{document}